\documentclass[review]{elsarticle}

\usepackage{graphicx}
\usepackage{amsmath}
\usepackage{amsfonts}
\usepackage{algorithm, algorithmic}
\usepackage{subfigure}
\usepackage{booktabs}
\usepackage{geometry}

\begin{document}

\begin{frontmatter}

\title{Graph Classification Based on Skeleton and Component Features}

\author[1]{Xue Liu}
\author[2,3,4,5]{Wei Wei\corref{cor1}}\ead{weiw@buaa.edu.cn}
\author[2,3,5]{Xiangnan Feng}
\author[2]{Xiaobo Cao}
\author[1]{Dan Sun}

\address[1]{Beijing System Design Institute of Electro-Mechanic Engineering, Beijing, 100854, China}
\address[2]{School of Mathematical Sciences, Beihang University, Beijing, 100191, China}
\address[3]{Key Laboratory of Mathematics, Informatics and Behavioral Semantics, Ministry of Education, 100191, China}
\address[4]{Beijing Advanced Innovation Center for Big Data and Brain Computing, Beihang University, Beijing, 100191, China}
\address[5]{Peng Cheng Laboratory, Shenzhen, Guangdong, 518066, China}

\cortext[cor1]{Corresponding author}

\begin{abstract}

\end{abstract}


\end{frontmatter}


\section{Introduction}
Graph classification to distinguish the class labels of graphs in a dataset is an important task with practical applications in a large spectrum of fields (e.g., bioinformatics~\cite{2_lee2018graph}, social network analysis~\cite{3_2011Supervised_random_walks} and chemoinformatics~\cite{4_2010Fast}). In these areas, data can be usually represented as graphs with labels. For example, in bioinformatics, a protein molecule can be represented as a graph whose nodes corresponds to atoms, and edges signify there exits chemical bonds or not between atoms. The graphs are allocated with different labels based on having specific function or not. To make classification in this task, we usually make a common assumption that protein molecules with similar structure have similar functional properties.

More recently, there has been a surge of approaches that seek to learn representations or embeddings that encode features about the graphs and then make classification. The idea behind these learning approaches focuses on graph structure representation and learning a mapping that embeds nodes or entire (sub)graphs, into a low-dimensional vector. Most of these methods can be classified into two categories: (1) neural networks manners~\cite{8_niepert2016learning} that learn the large-scale structures of target graph, (2) kernel methods~\cite{5_Shervashidze2010Weisfeiler} that learn small-size structures of target graph. Different structures of graph imply dissimilar features.

Graph neural networks (GNNs)~\cite{6_2020A} use a recurrent network framework to transmit information from a calculated node to another new node until reaching a stop situation. Analogous to image-based convolutional neural networks (CNNs)~\cite{7_krizhevsky2012imagenet}, PATCHY-SAN (PSCN)~\cite{8_niepert2016learning} is motivated to operate on locally connected regions of the input to learn graph embeddings. Graph convolutional networks (GCNs)~\cite{9_kipf2016semi} operate directly on graph data using spectral filters to exploit local areas and then extract local meaningful features shared with the entire graph to get a large-size graph structure representation. The success of neural networks relies on enormous amount of data, and usually uses iterative calculations to spread information, so that the local information of the graph get coupled and is integrated into the overall embedding.

Subgraph isomorphism has been proven to be NP-complete, however graph isomorphism problem is in NP and has been neither proven NP-complete nor could be solved by a polynomial-time algorithm~\cite{10_hartmanis1982computers}. Graph kernels~\cite{5_Shervashidze2010Weisfeiler} differentiate two graphs by recursively decomposing them into substructures and defining a function on graph to make classification based on graphs similarity measures in an unsupervised way. They bridge the gap between graph data and a wide range of machine learning methods such as Support Vector Machines (SVM), regression, clustering and Principal Components Analysis (PCA), etc. Several different graph kernels approaches are usually divided into two classes: walk-based patterns and limited-size subgraph methods. In random walk pattern, graph kernels count matched random walks pairs between two graphs~\cite{11_kashima2003marginalized}. The shortest path kernels count pairs of shortest paths having the same beginning node and sink labels and the same length in two graphs~\cite{12_borgwardt2005shortest}. Graphlet kernels discuss graph isomorphism by counting the occurrences of all types of fixed size subgraphs~\cite{13_kondor2009graphlet}. Graphs are regarded similar if they share lots of common subgraphs. ~\cite{4_2010Fast} sets the source nodes of two graphs at a fixed distance from each other, and then finds subgraphs containing nodes up to a certain distance from the root and calculate the number of identical pairs of subgraphs. Weisfeiler-Lehman graph kernels are highly efficient kernels to express graph isomorphism based on comparing subtree-like patterns~\cite{5_Shervashidze2010Weisfeiler}.

\begin{figure*}[htbp]
\centering
\includegraphics[height=8cm,width=15cm]{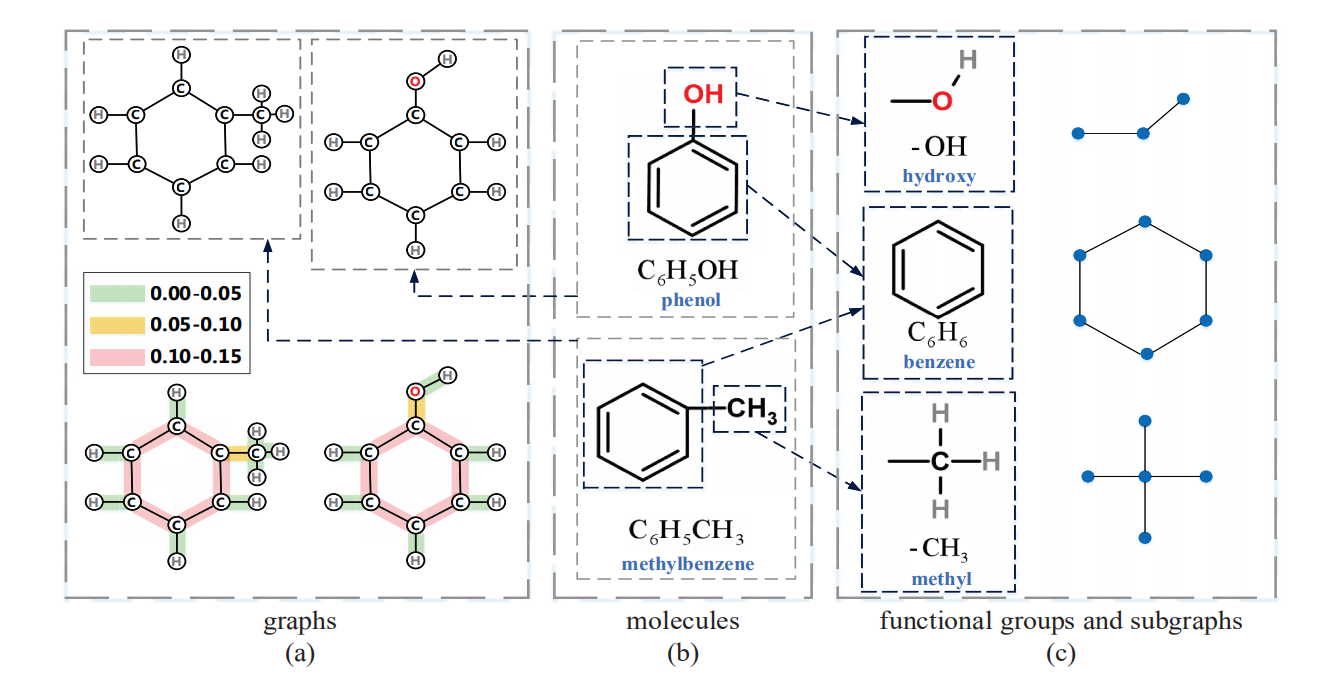}
\caption{
Overview of topological graphs (as shown in (a)), molecular chemical structures (as shown in (b)), their corresponding functional groups and subgraphs (as shown in (c)) of phenol ($\textrm{C}_{6}\textrm{H}_{5}\textrm{O}\textrm{H}$) and methylbenzene ($\textrm{C}_{6}\textrm{H}_{5}\textrm{C}\textrm{H}_{3}$). In (a), we do random walks with walk length $l = 4$ from each node on every topological graph of phenol and methylbenzene. The frequency $p$ of each edge covered in random walks is $p = \frac{m_{e}}{M}$, where $m_{e}$ is the frequence of edge $e$ being covered and $M$ is the total number of edges being covered. Edges with different frequencies are represented in different colors. Phenol and methylbenzene in (c) can be decomposed into three functional groups (hydroxy, benzene and methyl), which are regarded as 2-order subgraph (2-hop path), hexagon loop (6-order subgraph) and cross separately.
}
\label{Figure-1}
\end{figure*}

There are two critical limitations of graph kernels: (1) Many of them do not provide explicit embeddings, so that kernels are unusable for many proposed machine learning algorithms which operate on vector directly. (2) The substructures (i.e., random walks, subgraphs, etc.) need to be manually set priority, for instances, the length of random walk or the shape of subgraphs. This makes some substructures not in specified shape be ignored easily.

Anonymous Walk Embeddings (AWE) proposes a novel graph embedding method that relies on distribution of special random walks named anonymous random walks to get graph representations~\cite{14_ivanov2018anonymous}. In analogy to graph kernels and to avoid sparse distribution, AWE method uses random walks in anonymous manner to catch ``skeletons'' of the whole graph. ~\cite{15_micali2016reconstructing} has shown that anonymous walks are Markov processes from starting node. After adequate sampling, anonymous walks are capable to reconstruct original graph. Two graphs with similar distributions of anonymous walks are regarded topological similar~\cite{14_ivanov2018anonymous}.

However, AWE may ignore important subgraphs which actually determine graph properties because of low distribution. As illustrated in Figure \ref{Figure-1}, we take phenol ($\textrm{C}_{6}\textrm{H}_{5}\textrm{O}\textrm{H}$) and methylbenzene ($\textrm{C}_{6}\textrm{H}_{5}\textrm{C}\textrm{H}_{3}$) as simple examples. Each of them can be decomposed into two main parts, the main structure (benzene ring) as a skeleton, and functional groups (i.e., hydroxy ``$-\textrm{O}\textrm{H}$'' and methyl ``$-\textrm{C}\textrm{H}_{3}$''), and both of them determine the molecular properties and functions. Taking these two molecules as topological graphs and doing random walks on them, experiment shows that the frequency of benzene ring edges being covered is twice more than that of hydroxy or methyl edges being covered. In AWE, these two graphs are structure similar since random walks catch same main structures as skeletons. However, phenol and methylbenzene have quite different chemical properties determined by different functional groups, which are in low distributions. In addition, AWE lacks a hierarchical representation of the entire graph structure. Structures AWE gets have identical size (walk length), so that substructures whose size not equal to anonymous walks structures can not be represented efficiently.

\noindent\textbf{Our approach.} We design a novel data-driven framework named GraphCSC that represents entire graph in low-dimension vectors containing both main structure features and component information in a global view. Our methodology uses anonymous walks to represent skeleton information of a graph inspired by the success of AWE; to learn graph component representation, finding all different order subgraphs and checking subgraph isomorphism needs great computational complexity~\cite{16_nguyen2018learning}. Thus, our strategy is to embed graph using a distribution of special subgraphs, ie., frequent subgraphs. Frequent subgraphs are determined by a fixed threshold hyperparameter, which indicates what kinds of subgraphs can be regarded frequent. In frequent-based subgraphs, it not only contains the underlying semantics within an individual graph but also the relationships among graphs. To learn graph embedding, motivated by a novel supervised document method \textbf{PV-DBOW} (paragraph vector-distributed bag of words)~\cite{17_le2014distributed}, anonymous walks and frequent subgraphs in our model are treated as words and each graph is regarded as a document. Two graphs are similar in embedding space if their skeletons and components information are similar.

\noindent\textbf{Our contribution.} To the best of our knowledge, GraphCSC is a new framework that learns embeddings consisting both skeleton features and component information compared to other existing embedding approaches. GraphCSC actually studies the representation of graph structures from horizontal and vertical perspectives: in horizontal perspective, we attempt to characterize relatively high-order structures using anonymous random walks with same walk length, which determines the skeletons of graphs; in vertical perspective, our model focuses on what kinds of subgraphs with different sizes or shapes they have. We use a \textbf{NLP} (Natural Language Processing) training framework with skeletons and components together as inputs to learn graph representation with combined information. Through empirical evaluation on multiple real-world datasets, experiments show that our model is competitive than various established baselines.

\section{PROBLEM STATEMENT}
\noindent\textbf{Definition 1 (Graph Classification)}. Given a set of graphs with labels, $\mathbb{G}$ = $\{\emph{G}_1, \ldots, \emph{G}_N\}$, the goal is to learn a function $\varphi: \mathbb{G} \rightarrow \mathbb{L}$, where $\mathbb{G}$ is the input space of graphs and $\mathbb{L}$ is the set of graph labels.  Each weighted graph \emph{G}$_i$ with label $l_i$, $i = 1, \ldots, N$, is a tuple $G_i$ = $(V_i, E_i, \Omega_i)$, where $V_i$ is the set of $n_i$ vertices from $G_{i}$, $E_i \subseteq V_i\times V_i$ is the set of edges from $G_{i}$, and $\Omega_i$ is the set of edge weights from $G_{i}$.

\noindent\textbf{Definition 2 (Graph Embedding)}. Given a set of graphs $\mathbb{G}$ = $\{G_1, \ldots, G_N\}$, the goal is to learn a graph embedding matrix $\boldsymbol{X}_{N \times d}$, where each $i$-th row is a $d$-dimensions vector of graph $G_{i}$  which is learned by mapping $\psi: \mathbb{G} \rightarrow \mathbb{R}^{1\times d}$. Graph embeddings capture the graph similarity between $G_i$ and $G_j$ in the sense that vector $\psi(G_i)$ and $\psi(G_j)$ are close in embedding space.

\section{BACKGROUND}
We will leverage two techniques \textbf{Skipgram} and \textbf{PV-DBOW} which have achieved success in \textbf{NLP} to learn graph representation. Before we propose our approach, we review these powerful models and state them as background of our model.

\subsection{\textbf{word2vec} and Skipgram}
How to get continuous-valued word vector representations is a core task in \textbf{NLP} applications. \textbf{word2vec}~\cite{18_2013Distributed} uses \textbf{Skipgram} to learn low-dimension embeddings of words that capture rich semantic relationships between words. \textbf{Skipgram} maps words contained in similar sentences to ``near'' positions in embedding space, i.e., their representation vectors are similar.

Given the target word $w_{t}$ from vocabulary set $V$ and a sequence of words $w_{1},\ldots,w_{t},\ldots,w_{T}$, a context $w_{t-c},\ldots,w_{t},\ldots,w_{t+c}$ is defined as a fixed number of words surrounding $w_{t}$ within a window $c$. \textbf{Skipgram} maximizes the co-occurrence probability among words that appear in context:
\begin{equation}\label{(1)}
  \sum\limits_{t=1}^{T}\log\bold{Pr}(w_{t-c},\ldots,w_{t+c}|w_{t}).
\end{equation}
The conditional probability $\bold{Pr}(w_{t-c},\ldots,w_{t+c}|w_{t})$ is approximated under the following independence assumption:
\begin{equation}\label{(2)}
  \bold{Pr}(w_{t-c},\ldots,w_{t+c}|w_{t}) = \prod\limits_{j=t-c, j\neq{t}}^{t+c}\bold{Pr}(w_{j}|w_{t}).
\end{equation}

\subsection{Softmax and negative sampling}
To learn such a posterior distribution $\bold{Pr}(w_{t+j}|w_{t})$, conventional classifier such as logistic regression requires vast computational resources since the number of labels equals to vocabulary size $|V|$. To avoid heavy calculation, conditional probability distribution $\bold{Pr}(w_{j}|w_{t})$ is defined by a Hierarchical Softmax~\cite{19_mnih2009scalable}:
\begin{equation}\label{(3)}
  \bold{Pr}(w_{j}|w_{t})=\frac{\exp(\boldsymbol{w}_{t}\cdot\boldsymbol{w}_{j})}
  {\sum\limits_{i=1}^{|V|}\exp(\boldsymbol{w}_{t}\cdot\boldsymbol{w}_{i})},
\end{equation}
where $\boldsymbol{w}_{t}$ and $\boldsymbol{w}_{t+j}$ are embedding vectors of word $w_{t}$ and $w_{t+j}$. Thus the co-occurrence probability (\ref{(1)}) is written as:
\begin{equation}\label{(4)}
  \sum\limits_{t=1}^{T}\log\prod\limits_{j=t-c, j\neq{t}}^{t+c}\frac{\exp(\boldsymbol{w}_{t}\cdot\boldsymbol{w}_{j})}
  {\sum\limits_{i=1}^{|V|}\exp(\boldsymbol{w}_{t}\cdot\boldsymbol{w}_{i})}.
\end{equation}

To speed up the training process of \textbf{Skipgram}, negative sampling~\cite{20_rousseau2015text} method random samples a small set of words as negative samples which are not involved in context. Then only target word and negative samples are updated instead of the whole words from vocabulary set $V$ in the process of iteration training. This strategy would be efficient especially for situations when tasks face huge computational pressure.

\subsection{\textbf{doc2vec} and \textbf{PV-DBOW}}
In analogy to \textbf{word2vec}, \textbf{doc2vec}~\cite{17_le2014distributed} uses \textbf{PV-DBOW} to learn representations of arbitrary size document in a document set.
More specifically, given a document set $\mathbb{D} = \{D_{1},\ldots,D_{N}\}$ with a set of words $V=\{w_{1},\ldots,w_{|V|}\}$ and the target document $D_{t}\in \mathbb{D}$ contains a sequence of words $\{w_{1},\ldots,w_{l}\}$, the goal is to learn a low-dimension vector $\textbf{D}_{t}$ of document $D_{t}$ by maximize the following log probability of words $w_{1},\ldots,w_{l}$ contained in $D_{t}$:

\begin{equation}\label{(5)}
  \sum\limits_{i=1}^{l}\log\bold{Pr}(w_{j}|D_{t}).
\end{equation}

The conditional probability $\bold{Pr}(w_{j}|D_{t})$ above is defined as:
\begin{equation}\label{(6)}
  \bold{Pr}(w_{j}|D_{t})=
  \frac{\exp(\boldsymbol{D}_{t}\cdot\boldsymbol{w}_{j})}
  {\sum\limits_{i=1}^{|V|}\exp(\boldsymbol{D}_{t}\cdot\boldsymbol{w}_{i})},
\end{equation}
where $\boldsymbol{D}_{t}$ and $\boldsymbol{w}_{j}$ are corresponding representation of $D_{t}$ and $w_{j}$, $|V|$ is the number of all words across all documents in $\mathbb{D}$. The log probability (\ref{(5)}) could be approximated efficiently using negative sampling.

\section{PROPOSED MODEL}
Our proposed model GraphCSC has two main modules, skeleton module and component module as summarized in Figure \ref{Figure-5}. Skeleton module (in \textbf{Algorithm} \ref{Algorithm-1}) and component module (in \textbf{Algorithm} \ref{Algorithm-2}) mine corresponding substructures separately but synchronously. Finally, GraphCSC integrates these two modules and optimizes overall loss function by gradient descent.

\subsection{Skeleton Module}
\subsubsection{Anonymous Random Walks}
\noindent\textbf{Definition 3 (Random Walk).} In graph $G$, a random walk $w$ is defined as a finite sequence $(v_{0}, v_{1},\dots, v_{l})$ with length $l$, where $v_{0}$ is the root node and node $v_{i+1}$ is sampled independently among the neighbors of node $v_{i}$.

Random walks are regarded as Markov processes, recently anonymous random walks have been proven to be capable to learn graphs structural properties and reconstruct graphs with full descriptions of every node's state appeared in the random walk process in its own label space instead of global label space~\cite{15_micali2016reconstructing}. This makes random walks in anonymous experiment more flexible and compact.

\noindent\textbf{Definition 4 (Position Function)}. Let $w = (v_0, v_1, \ldots, v_l)$ be a random walk with length \emph{l} on graph. The position function \emph{f} is defined as $f(v_i) = |(v_{0}, \ldots, v_{i'})|$, where $i’$ is the smallest integer such that $v_{i'} = v_{i}$.

\noindent\textbf{Definition 5 (Anonymous Random Walk)}. Let $w$ = $(v_0, v_1, \ldots, v_l)$ be a random walk with length \emph{l} on graph \emph{G}$_i$. The corresponding anonymous random walk is defined as a sequence of integers $a = g(w) = (f(v_{0}), f(v_{1}), \ldots, f(v_{l}))$, where \emph{f} is the position function.

\begin{figure}[htbp]
\centering
\includegraphics[height=9cm,width=8cm]{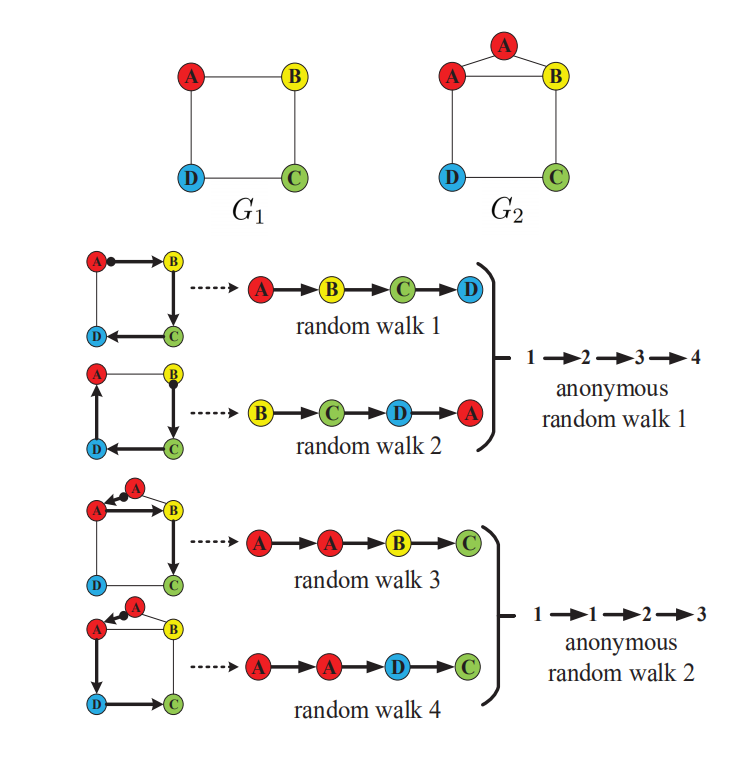}
\caption{
We sample four random walks with length $3$ (marked in different colors) in $G_{1}$, $G_{2}$ and illustrate their relation with corresponding anonymous random walks.
}
\label{Figure-2}
\end{figure}

\noindent\textbf{Example.} Random walks $(A, B, C, D)$, $(B, C, D, A)$, $(A, A, B, C)$ and $(A, A, D, C)$ sampled from $G_{1}$ and $G_{2}$ in Figure \ref{Figure-2} are completely different. However, in anonymous random walks view, the new label for each node is redefined as the position of the first occurrence of node with same label in the random walk sequence. This makes the initial four different random walks be changed into two anonymous random walks, $(1, 2, 3, 4)$ and $(1, 1, 2, 3)$.

\begin{figure}[htbp]
\centering
\includegraphics[height=4.7cm,width=9cm]{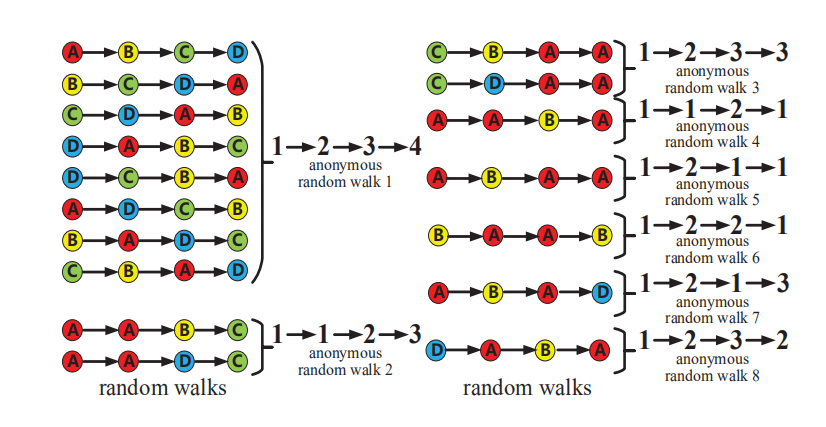}
\caption{
All kinds of different random walks with length $3$ in $G_{1}$ and $G_{2}$ from Figure \ref{Figure-2} and their corresponding anonymous random walks.
}
\label{Figure3}
\end{figure}

As shown in Figure \ref{Figure3}, all different random walks with length $3$ in $G_{1}$ and $G_{2}$ from Figure \ref{Figure-2} will be converted into only $8$ anonymous random walks. Random walks record the labels information of nodes traveled, under sufficient samples, they can accurately capture structure traits and reconstructure the original graph. But over precise information will not be efficient to capture graph structure features because of sparse distribution over all random walks. Anonymous random walks translate random walks into a sequence of integers recording first appearing positions.

\subsubsection{Skeleton}\
We draw independently a set of $\xi$ random walks $W_{G_i} = \{w_{1}, \ldots, w_{\xi}\}$ with length $l$ on $G_{i}$ in graphs set $\mathbb{G} = \{G_1, \ldots, G_i, \ldots, G_N\}$, and calculate its $\mu$ anonymous random walks $A_{G_i} = \{a_{1}, \ldots a_{\mu}\}$.

For a large graph $G_{i}$, to count all possible anonymous random walks needs vast computational resources. However, sampling $\zeta$ random walks with length $l$ to approximate actual distribution of anonymous random walks, the overall computing running time will be $\emph{O}(\zeta l)$ [16]. The relation of estimation of samples number $\zeta$ and the number of anonymous random walks $\lambda$ is determined by hyperparameters $\varepsilon$ and $\delta$~\cite{26_shervashidze2009efficient}:
\begin{equation}\label{7}
  \zeta = [\frac{2}{\varepsilon^{2}}(log(2^{\lambda}-2)-log(\delta))], \varepsilon > 0, 0\leq \delta \leq1.
\end{equation}

Next we tend to leverage vectors to represent whether some specific anonymous random walks contained in a graph or not.

\noindent\textbf{Definition 6 (Skeleton)}. Given graphs set $\mathbb{G} = \{G_1, \ldots, G_i, \ldots, G_N\}$, we sample random walks with length $l$ on each graph in $\mathbb{G}$ and integrate them as set $W = \{w_{1}, \ldots, w_{m}, \ldots, w_{\xi}\}$. The correspond anonymous random walks set $A = \{a_{1}, \ldots, a_{s}, \ldots, a_{\mu}\}$ has $\mu$ unequal elements. The \textbf{skeleton} of graph $G_{i}$ is defined as an $1 \times \mu$ shape vector $\boldsymbol{s}_{G_i}=[\xi_{s}]_{1 \times \mu}$, where
\begin{equation}\label{0}
     \quad\xi_{s}=\left\{
     \begin{aligned}
      1&,\ if\ a_{s} \in G_{i} \\
      0&,\ else
      \end{aligned}
      \right. , s=1,\ldots,\mu.
\end{equation}

\begin{algorithm}
	\renewcommand{\algorithmicrequire}{\textbf{Input:}}
	\renewcommand{\algorithmicensure}{\textbf{Output:}}
	\caption{Skeleton Module}
	\label{Algorithm-1}
	\begin{algorithmic}[1]
		\REQUIRE $\mathbb{G}$ = $\{G_1, \ldots, G_N\}$: graphs set; $l$: random walk length; $T$: random walk times.
		\ENSURE $A$: anonymous random walks set; $\boldsymbol{s}_{G_{i}}$, $i=1,\ldots,N$: \textbf{skeleton} for each graph $G_{i}$ in $\mathbb{G}$.
        \STATE \% \emph{Mining Anonymous Walks}
        \STATE $A=\emptyset$
        \FOR {each $G_{i}$ in $\mathbb{G}$}
            \STATE $A_{G_{i}}=\emptyset$
            \FOR{each node $v_j$ in $V(G_{i})$}
                \FOR{$k=1$; $k<T$; $k++$}
                    \STATE $w=RandomWalk(G_{i},v_{j}, l)$
                    \STATE $a = g(w)$
                    \IF{$a$ not in $A_{G_{i}}$} \STATE $A_{G_{i}}=A_{G_{i}} \cup\{a\}$
                    \ENDIF
                \ENDFOR
            \ENDFOR
        \ENDFOR
        \STATE $A=\bigcup\limits_{i=1}^{N}A_{G_{i}}$

        \STATE\% \emph{Getting} \textbf{skeleton}\emph{s}
        \FOR {each $G_{i}$ in $\mathbb{G}$}
            \STATE initialize $\boldsymbol{s}_{G_{i}}=[s_{k}]_{1 \times |A|}$, $s_{k}=0$
            \FOR {each anonymous random walk $a_{k}$ in $A$}
                \IF {$a_{k}$ in $A_{i}$}
                    \STATE{$s_{k}=1$}
                \ENDIF
            \ENDFOR
        \ENDFOR
        \STATE \textbf{return} $A$; $\boldsymbol{s}_{G_{i}}$, $i=1,\ldots,N$
	\end{algorithmic}
\end{algorithm}

\subsection{Component Module}
Component module uses a more flexible approach to capture similarity between two graphs by frequent patterns, for example, itemsets, subsequences, or subgraphs, which appear in a data set with frequency no less than a user-specified minimum support ($min\_sup$) threshold. Component module basically includes three steps: (1) mining frequent subgraphs,  (2) selecting features, (3) learning component features.

\subsubsection{Frequent Subgraph}\

\noindent\textbf{Definition 7 (Subgraph)}. A graph $sg$ is called a subgraph of a graph $G$ if nodes satisfy \textbf{V($sg$)} $\subseteq$ \textbf{V($G$)} and edges satisfy \textbf{E($sg$)} $\subseteq$ \textbf{E($G$)}.

\begin{figure}[htbp]
\centering
\includegraphics[height=6cm,width=7cm]{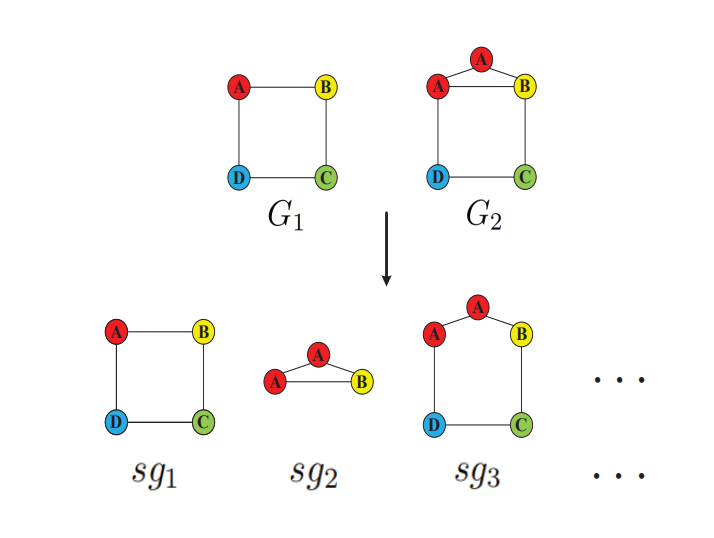}
\caption{Different kinds of subgraphs mined from $G_{1}$ and $G_{2}$ in Figure \ref{Figure-2}.}
\label{Figure-4}
\end{figure}

Two graphs are similar if their subgraphs are similar since subgraphs have been recognized as fundamental units and are building blocks for complex networks~\cite{29_milo2002network}. But usually only a fraction of the large amount number of subgraphs are actually relevant to data mining problems. Our methodology focus on mining subgraphs of different proportion in graph set, and then embed these distribution features into graph component representation.

\noindent\textbf{Definition 8 (Frequent Subgraph)}.
Let $sg$ be a subgraph in subgraphs set $SG$ mined from $\mathbb{G} = \{G_1, \ldots, G_i, \ldots, G_N\}$, and will be called frequent subgraph $fsg$ if  $\frac{|\{G_{i}\in\mathbb{G}|sg \in G_{i}\}|}{|\mathbb{G}|}$ $\geq \theta$, where $\theta$ is the $min\_sup$ threshold, $0 \leq \theta \leq 1$, and $\frac{|\{G_{i}\in\mathbb{G}|sg \in G_{i}\}|}{|\mathbb{G}|}$ is called the $support$ of $G_{i}$.

\noindent\textbf{Example.} We still take $G_{1}$, $G_{2}$ in Figure \ref{Figure-2} and subgraphs $sg_{1}$, $sg_{2}$, $sg_{3}$ in Figure \ref{Figure-4} as examples. If we set $min\_sup$ threshold $\theta=1$, only subgraphs $sg_{1}$ can be regarded as frequent subgraph. However, if we set $min\_sup$ threshold $\theta=0.5$, all of $sg_{1}$, $sg_{2}$ and $sg_{3}$ will be regarded frequent.

Frequent pattern, as a form of non-linear feature combinations over the set of different subgraphs, has higher discriminative power than that of single kind of subgraph because they capture more underlying semantics of the data. The key point is to specify the hyperparameter $min\_sup$ threshold $\theta$ used in model in frequent pattern, and we will study the influence of $\theta$ for machine learning tasks in \textbf{Section \ref{experiments}}. If an infrequent feature is used, the model cannot generalize well to the test data since it is built based on statistically minor observations, hence the discriminative power of low-support features will be limited~\cite{21_Hong2008Towards}.

\begin{figure*}[htb]
\centering
\includegraphics[height=8cm,width=15cm]{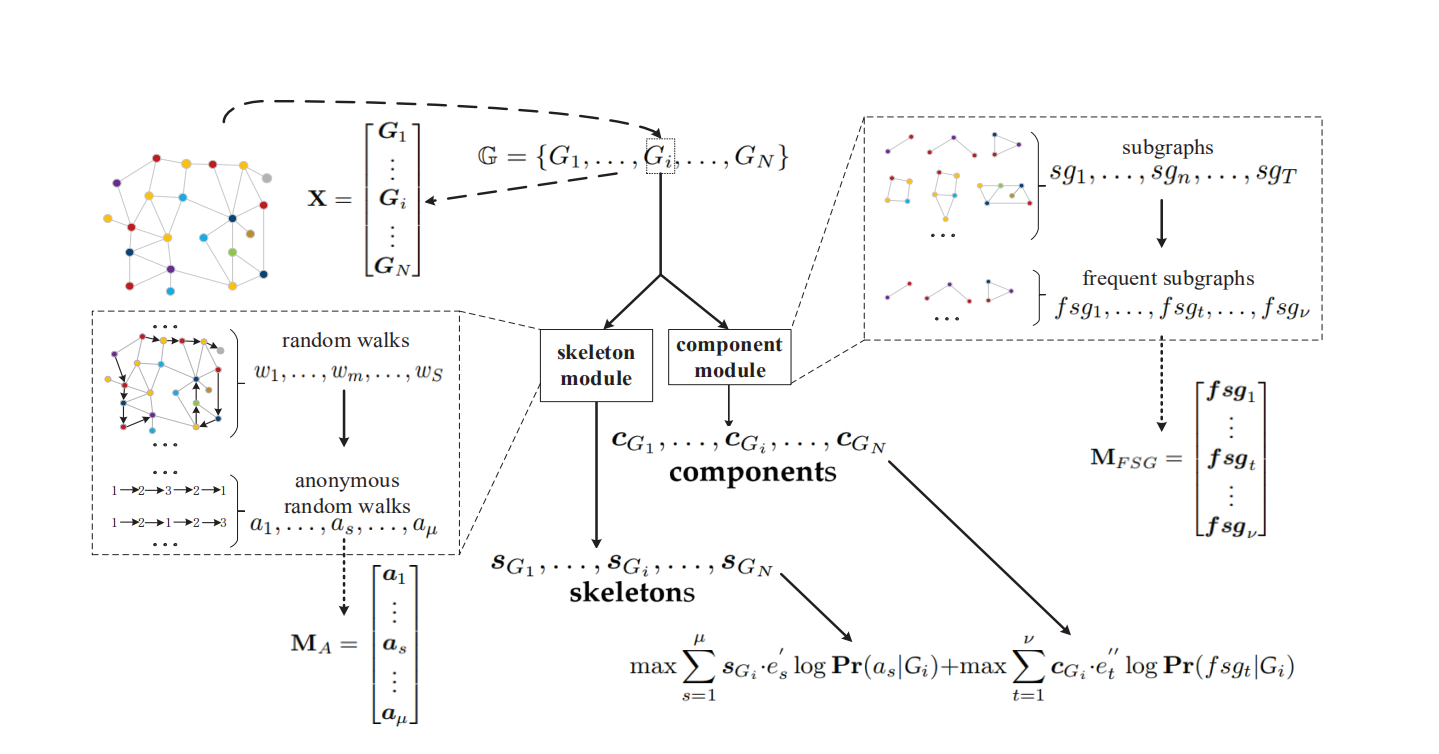}
\caption{
Overview of GraphCSC which can be decomposed into two main parts, the skeleton module and component module. Skeleton module outputs anonymous random walks matrix $\bold{M}_{A}$ in $\mu \times d$ size and \textbf{skeleton}s for graphs from $\mathbb{G}$. Meanwhile, component module gets frequent subgraphs matrix $\bold{M}_{FSG}$ in $\nu \times d$ size and \textbf{componnet}s for graphs from $\mathbb{G}$. Finally, GraphCSC integrates these two modules and yields global optimization object (\ref{14}).
}
\label{Figure-5}
\end{figure*}

\subsubsection{Component}\
To mine all frequent subgraphs will face two challenges: (1) The time complexity $O(2^{|V|})$ which is dominated by the overall nodes number of each graph, makes it computational infeasible to find all subgraphs. (2) In analogue of graph isomorphism checking in which two graphs have different sizes, subgraph isomorphism checking has been proven to be NP-complete~\cite{22_zaki2014data}. A slightly less restrictive measure of similarity can be deﬁned based on the size of the largest common subgraph in two graphs, but unfortunately the problem of ﬁnding the largest common subgraph of two graphs is NP-complete as well~\cite{22_zaki2014data}.

To tackle with such two challenges above, \textbf{gSpan}~\cite{31-gSpan-yan2002gspan} algorithm builds a new lexicographic order among graphs, and maps each graph to a unique minimum DFS-code as its canonical label. Based on this lexicographic order among DFS-code, \textbf{gSpan} adopts the depth-first search strategy to discover frequent subgraphs efficiently. Repeating this procedure until either when the $support$ of a graph is less than $min\_sup$ threshold $\theta$, or its code is not a minimum code, which means this graph and all its descendants have been generated and discovered before. Now we give the definition of \textbf{component} to indicate what frequent subgraphs mined by \textbf{gSpan} graph $G_{i}$ has.

\noindent\textbf{Definition 9 (Component)}. Given a set of graphs $\mathbb{G} = \{G_{1},\ldots, G_{i}, \ldots, G_{N}\}$ and $min\_sup$ threshold $\theta$, all different frequent subgraphs obtained by \textbf{gSpan} are contained in set $FSG$ = $\{fsg_1, \ldots, fsg_t, \ldots, fsg_{\nu}\}$. The \textbf{component} of $G_i$  is defined as an $1 \times \nu$ shape vector $\boldsymbol{c}_{G_i}=[\eta_{t}]_{1 \times \nu}$, where
\begin{equation}\label{0}
     \quad\\ \eta_{t}=\left\{
     \begin{aligned}
      1&,\ if\ fsg_{t} \in G_{i} \\
      0&,\ else
      \end{aligned}
      \right. , t=1,\ldots,\nu.
\end{equation}

\begin{algorithm}[htbp]
	\renewcommand{\algorithmicrequire}{\textbf{Input:}}
	\renewcommand{\algorithmicensure}{\textbf{Output:}}
	\caption{Component Module}
	\label{Algorithm-2}
	\begin{algorithmic}[1]
		\REQUIRE $\mathbb{G}$ = $\{G_1, \ldots, G_N\}$: graphs set; $\theta$: $min\_sup$ threshold.
		\ENSURE $FSG$: frequent subgraphs set; $\boldsymbol{c}_{G_{i}}$, $i=1,\ldots,N$: \textbf{component} for each graph $G_{i}$ in $\mathbb{G}$

        \STATE \% \emph{Mining Frequent Subgraphs}
		\STATE $FSG=\textbf{gSpan}(\mathbb{G}, \theta)$
		\FOR {each $G_{i}$ in $\mathbb{G}$}
            \STATE $FSG_{G_{i}}=\emptyset$
            \FOR {each frequent subgraph $fsg_{j}$ in $FSG$}
                \IF{$fsg_{j}$ in $G_{i}$}
                    \STATE $FSG_{G_{i}}=FSG_{G_{i}}\cup \{fsg_{j}\}$
                \ENDIF
            \ENDFOR
        \ENDFOR

        \STATE \% \emph{Getting} \textbf{component}\emph{s}
        \FOR {each $G_{i}$ in $\mathbb{G}$}
            \STATE initialize $\boldsymbol{c}_{G_{i}}=[c_{k}]_{1 \times |FSG|}$, $c_{k}=0$
            \FOR {each frequent subgraph $fsg_{k}$ in $FSG$}
                \IF {$fsg_{k}$ in $FSG_{i}$}
                    \STATE{$s_{k}=1$}
                \ENDIF
            \ENDFOR
        \ENDFOR
        \STATE \textbf{return} $FSG$; $\boldsymbol{c}_{G_{i}}$, $i=1,\ldots,N$
	\end{algorithmic}
\end{algorithm}

\subsection{TRAINING}
For graph data set $\mathbb{G} = \{G_{1},\ldots, G_{i}, \ldots, G_{N}\}$, supposed that all anonymous random walks with length $l$ mined are in $A =\{a_{1},\ldots,a_{\mu}\}$ and all frequent subgraphs mined under $min\_sup$ threshold $\theta$ are in $FSG=\{fsg_{1},\ldots,fsg_{\nu}\}$ , we leverage \textbf{PV-DBOW} to learn graphs embedding matrix $\bold{X}$ whose each row is an embedding for each graph, anonymous random walks matrix $\bold{M}_{A}$ whose each row is an embedding for each anonymous random walk and frequent subgraphs matrix $\bold{M}_{FSG}$ whose each row is an embedding for each frequent subgraph. Vectors $\boldsymbol{s}_{G_i}$ and $\boldsymbol{c}_{G_{i}}$ are corresponding \textbf{skeleton} and \textbf{component} for graph $G_i$, $i=1,\ldots,N$.

Graphs embedding matrix $\bold{X}$ has $N \times d$ size, matrix $\bold{M}_{A}$ is in $\mu \times d$ size and matrix $\bold{M}_{FSG}$ is in $\nu \times d$ size, where $d$ is embedding dimension. \textbf{PV-DBOW} treats graph data set $\mathbb{G}$ as a documents set, each graph $G_i$ in $\mathbb{G}$ as a document and each substructure mined in $G_i$ as a word contained in a document.

To begin with, we focus on anonymous random walks and seek to optimize the following objective function, which maximizes the log-probability of predicting anonymous random walks that appear in graph $G_i$:
\begin{equation}\label{equation_10}
  \max\sum\limits_{s=1}^{\mu}\boldsymbol{s}_{G_i}\cdot e^{'}_{s}\log\bold{Pr}(a_{s}|\emph{G}_{i}),
\end{equation}
where $e_{s}^{'}$ is an $1 \times \mu$ binary vector whose $s$-th column element is $1$ and $j$-th column element equals to $0$ if $j\neq s$.

The probability $\bold{Pr}(a_{s}|\emph{G}_{i})$ is defined as a softmax unit parametrized by a dot product of $\boldsymbol{a}_{s}$ and $\boldsymbol{G}_{i}$ which are embedding vectors of $a_{s}$ and $G_{i}$:
\begin{equation}\label{equation_11}
  \bold{Pr}(a_{s}|\emph{G}_{i}) = \frac{\exp(\boldsymbol{a}_{s} \cdot \boldsymbol{G}_{i})}{\sum\limits_{p=1}^{\mu}\exp(\boldsymbol{a}_{p} \cdot \boldsymbol{G}_{i})}.
\end{equation}

Next, we use same method to maximize the log-probability of predicting the frequent subgraphs that appear in graph $G_i$:
\begin{equation}\label{equation_12}
  \max\sum\limits_{t=1}^{\nu}\boldsymbol{c}_{G_i}\cdot e^{''}_{t}\log\bold{Pr}(fsg_{t}|\emph{G}_{i}),
\end{equation}
where $e_{t}^{''}$ is an $1 \times \nu$ binary vector whose $t$-th column element is $1$ and $j$-th column element equals to $0$ if $j\neq t$.

The probability $\bold{Pr}(fsg_{t}|\emph{G}_{i})$ is defined as a softmax function parametrized by a dot product of $\boldsymbol{fsg}_{t}$ and $\boldsymbol{G}_{i}$ which are embedding vectors of subgraph $fsg_{t}$ and $G_{i}$:
\begin{equation}\label{13}
  \bold{Pr}(fsg_{t}|\emph{G}_{i}) = \frac{\exp(\boldsymbol{fsg}_{t} \cdot \boldsymbol{G}_{i})}{\sum\limits_{q=1}^{\nu}\exp(\boldsymbol{fsg}_{q} \cdot \boldsymbol{G}_{i})}.
\end{equation}

To get graph representation satisfies (\ref{equation_10}) and (\ref{equation_12}), the global optimization object is:
\begin{equation}\label{14}
  \max\sum\limits_{s=1}^{\mu}\boldsymbol{s}_{G_i}\cdot e^{'}_{s}\log\bold{Pr}(a_{s}|\emph{G}_{i})+
  \max\sum\limits_{t=1}^{\nu}\boldsymbol{c}_{G_i}\cdot e^{''}_{t}\log\bold{Pr}(fsg_{t}|\emph{G}_{i}).
\end{equation}

We yield its loss function as following:
\begin{equation}\label{(15)}
\begin{split}
   \bold{L} = & \sum\limits_{s=1}^{\mu} \boldsymbol{s}_{G_i}\cdot e^{'}_{s}\log \sigma(\boldsymbol{a}_{s} \cdot \boldsymbol{G}_{i}) + \sum\limits_{t=1}^{\nu}\boldsymbol{c}_{G_i}\cdot e^{''}_{t} \log\sigma(\boldsymbol{fsg}_{t} \cdot \boldsymbol{G}_{i})\\
     & \sum\limits_{p=1}^{\mu}\log(\sigma(-\boldsymbol{a}_{p} \cdot \boldsymbol{G}_{i}))
   +  \sum\limits_{q=1}^{\nu}\log(\sigma(-\boldsymbol{fsg}_{q} \cdot \boldsymbol{G}_{i})),
\end{split}
\end{equation}
where $\sigma(x) = \frac{1}{1 + exp(-x)}$ is sigmoid function.

The last two items in equation (\ref{(15)}) which sum over all anonymous random walks and subgraphs directly, are too expensive since $\mu$ and $\nu$ usually tend to be very large. Hence we proceed with an approximation by negative sampling to make the optimization problem tractable. The normalization terms from the softmax are replaced by $K_{1}$ anonymous random walks negative samples $\{{a'_{1},\ldots,a'_{K_{1}}}\}$ from $A$ but not contained in $G_{i}$ and $K_{2}$ frequent subgraphs negative samples $\{{fsg'_{1},\ldots,fsg'_{K_{2}}}\}$ from $F$ but not contained in $G_{i}$. Thus equation (\ref{(15)}) can be rewritten as:
\begin{equation}\label{16}
\begin{split}
   \bold{L} = & \sum\limits_{s=1}^{\mu} \boldsymbol{s}_{G_i}\cdot e^{'}_{s}\log \sigma(\boldsymbol{a}_{s} \cdot \boldsymbol{G}_{i}) + \sum\limits_{t=1}^{\nu}\boldsymbol{c}_{G_i}\cdot e^{''}_{t} \log\sigma(\boldsymbol{fsg}_{t} \cdot \boldsymbol{G}_{i}) \\
     & \sum\limits_{p=0}^{K_{1}}log(\sigma(-\boldsymbol{a'}_{p} \cdot \boldsymbol{G}_{i}))
   +  \sum\limits_{q=0}^{K_{2}}log(\sigma(-\boldsymbol{fsg'}_{q} \cdot \boldsymbol{G}_{i})),
\end{split}
\end{equation}
where $\boldsymbol{a'}_{p}$ and $\boldsymbol{fsg'}_{q}$ are the embedding vectors of samples $a'_{p}$ and $fsg'_{q}$ respectively, and $a'_{p}$ belongs to $G_{i}$ when $p=0$, $fsg'_{q}$ is in $G_{i}$ when $q=0$.

Finally, we optimize loss function (\ref{16}) with stochastic gradient descent and update $\boldsymbol{G}_{i}$. After the learning process finishes, two graphs are near in embedding space if they have similar skeleton and component, and we summarize this training process in \textbf{Algorithm} \ref{Algorithm-3}.

\begin{algorithm}
	\renewcommand{\algorithmicrequire}{\textbf{Input:}}
	\renewcommand{\algorithmicensure}{\textbf{Output:}}
	\caption{Training}
	\label{Algorithm-3}
	\begin{algorithmic}[1]
		\REQUIRE $\mathbb{G}$ = $\{G_1, \ldots, G_N\}$: graphs set; $A$: anonymous random walks set; $\boldsymbol{s}_{G_{i}}$, $i=1,\ldots,N$: \textbf{skeleton} for each graph $G_{i}$ in $\mathbb{G}$; $FSG$: frequent subgraphs set; $\boldsymbol{c}_{G_{i}}$, $i=1,\ldots,N$: \textbf{component} for each graph $G_{i}$ in $\mathbb{G}$; $e^{'}_{1},\dots,e^{'}_{|A|}$: $|A|$ binary vectors for \textbf{skeleton}s; $e^{''}_{1},\dots,e^{''}_{|FSG|}$: $|FSG|$ binary vectors for \textbf{component}s; $d$: embedding dimension; $\alpha$: learning rate.
		\ENSURE $\bold{X}^{N\times d}$: graphs representation matrix

        \STATE initialize $\bold{X}=[x_{ij}]_{N\times d}$, $x_{ij}\sim\emph{\textbf{N}}(0,0.001)$

    	\FOR {each anonymous walk $a_{s}$ in $A$}
            \FOR{each frequent subgraph $fsg_{t}$ in $FSG$}
                \STATE $\bold{L}(\bold{X})=-\boldsymbol{s}_{G_{i}}\cdot e^{'}_{s} \log\bold{Pr}(a_{s}|G_{i})- \boldsymbol{c}_{G_{i}}\cdot e^{''}_{t} \log\bold{Pr}(fsg_{t}|G_{i})$
                \STATE $\bold{X}=\bold{X}-\alpha\frac{\partial \bold{L}(\bold{X})}{\bold{X}}$
            \ENDFOR
        \ENDFOR
        \STATE \textbf{return} $\bold{X}$
	\end{algorithmic}
\end{algorithm}

\section{EXPERIMENTS}\label{experiments}
In this section, to quantitatively evaluate classification capability of our model, we conduct extensive experiments on a variety of widely-used datasets to compare with several state-of-the-art baselines.

\subsection{DATASETS}
 We evaluate our proposed method on binary classification task using seven real-world graph datasets whose statictics are summarized in Table \ref{table-1-dataset}. MUTAG~\cite{23_1991Structure} is a dataset of aromatic and heteroaromatic nitro compounds labeled according to whether or not they have a mutagenic effect on bacteria. PROTEINS~\cite{1_Borgwardt2005Protein} is a set of proteins graphs where nodes represent secondary structure elements and edges indicate neighborhood in the amino-acid sequence or in 3-dimension space. ENZYMES~\cite{1_Borgwardt2005Protein} consists of protein tertiary structures obtained from the BRENDA enzyme database. DD ~\cite{24_2003Distinguishing} is a dataset of protein structures where nodes represent amino acids and edges indicate spatial closeness, which are classiﬁed into enzymes or non-enzymes. PTC-MR~\cite{25_Helma2001The} consists of graph representations of chemical molecules  labeled according to carcinogenicity on rodents. NCI-1, NCI-109~\cite{2_lee2018graph} are datasets of chemical compounds divided by the anti-cancer property (active or negative). These datasets have been made publicly available by the National Cancer Institute (NCI).

\begin{table}
\centering
\caption{Statistics of the benchmark graph datasets. The columns are: Name of dataset, Number of graphs, Number of classes (maximum number of graphs in a class), Average number of nodes/edges.}

\begin{tabular}{lllll}
\toprule
Datasets & Graph $\sharp$ & Class $\sharp$ & Average Node $\sharp$ & Average Edge $\sharp$ \\
\midrule
MUTAG & 188 & 2 & 17.93 & 19.79\\
DD & 1178 & 2 & 284.32 & 715.66\\
PTC-MR & 344 & 2 & 14.29 & 14.69\\
NCI-1 & 4110 & 2 & 29.87 & 32.30\\
NCI-109 & 4127 & 2 & 29.68 & 32.13\\
PROTEINS & 1113 & 2 & 39.06 & 72.82\\
ENZYMES & 600 & 6 & 32.63 & 64.14\\
\bottomrule
\end{tabular}
\label{table-1-dataset}
\end{table}


\subsection{BASELINES}
In order to demonstrate the effectiveness of our proposed approach, we compare it with several baseline methods, all of which utilize the entire graph for feature extraction. These competitors can be categorized into four main groups:

\begin{itemize}
  \item \textbf{Graph kernels based methods:} The shortest path (SP)~\cite{12_borgwardt2005shortest} kernel measures the similarity of a pair of graphs by comparing the distance of the shortest paths between nodes in the graphs. Graphlet kernel (GK)~\cite{26_shervashidze2009efficient} measures graph similarity by counting the number of different graphlets and Deep GK~\cite{27_yanardag2015deep} is deep graphlet kernel. Weisfeiler-Lehman kernel (WL)~\cite{5_Shervashidze2010Weisfeiler} uses subtree pattern to mine structure information and Deep WL~\cite{27_yanardag2015deep} is deep Weisfeiler-Lehman kernel.
  \item \textbf{Unsupervised graph embedding methods:} node2vec~\cite{8_niepert2016learning} is an unsupervised task agnostic method that learns entire graph embedding. It proposes every graph into a fixed size vector containing distributed representation of graph structures.

  \item \textbf{Supervised graph embedding methods:} PSCN~\cite{9_kipf2016semi} is a convolutional neural network algorithm which has achieved strong classiﬁcation accuracy in many datasets.
  \item \textbf{Unsupervised graph embedding methods:} AWE~\cite{14_ivanov2018anonymous} uses an anonymous random walks approach to embed entire graphs in an unsupervised manner.
\end{itemize}

\begin{table*}
\centering
\caption{Classiﬁcation accuracy (standard deviation) of our method GraphCSC and state-of-the-art baselines on benchmark datasets. The last two rows denote the $\theta$ values and the $dim$ values used by our method for each dataset; these values are determined in Fig.5 and Fig.6. “-” means the classiﬁcation accuracy (standard deviation) is not available in the original papers.
}
\resizebox{\textwidth}{!}{
\begin{tabular}{lllllllllllll}
\toprule
\textbf{Algorithm} & \textbf{MUTAG}  & \textbf{DD} & \textbf{PTC-MR} & \textbf{NCI-1} & \textbf{NCI-109} & \textbf{PROTEINS} & \textbf{ENZYMES} \\
\midrule
WL & 80.63 (3.07)  & 77.95 (0.70) & 56.97 (2.01) & 80.13 (0.50) & 80.22 (0.34)& 72.92 (0.56) & 53.15 (1.14)\\
Deep WL & 82.95 (1.96)  & - & 59.17 (1.56)  & 80.31 (0.46) & 80.32 (0.33)& 73.30 (0.82)  & 53.43 (0.91)\\
GK & 81.66 (2.11)  & 78.45 (0.26) & 57.26 (1.41)& 62.28 (0.29)& 62.60 (0.19)& 71.67 (0.55)  & 26.61 (0.99)\\
Deep GK & 82.66 (1.45)  & - & 57.32 (1.13)  & 62.48 (0.25) & 62.69 (0.23) & 71.68 (0.50)  & 27.08 (0.79)\\
graph2vec & 83.15 (9.25)  & 58.64 (0.01) & 60.17 (6.86) & 73.22 (1.81) & 74.26 (1.47) & 73.30 (2.05)  & 44.33 (0.09)\\
PSCN & 92.63 (4.21)  & 77.12 (2.41) & 60.00 (4.82) & 78.59 (1.89) & - & 75.89 (2.76)  & -\\
AWE & 87.87 (9.76)   & 71.51 (4.02) & 59.14 (1.83) & 62.72 (1.67) & 63.21 (1.42) & 70.01 (2.52)  & 35.77 (5.93)\\
\midrule
\textbf{GraphCSC} & 88.42 (6.47)  & 89.38 (2.73) & 64.04 (3.61) & 85.70 (4.29) & 85.46 (3.66) & 76.71 (3.06) & 57.21 (5.70)\\
$\theta$ & 0.15   & 0.25 & 0.75  & 0.20 & 0.20 & 0.60 & 0.55\\
dim & 128  & 8 & 128 & 16 & 16 & 32 & 128\\
\bottomrule
\end{tabular}}
\label{table-2-results}
\end{table*}

\begin{figure*}[htb]
\centering
\includegraphics[height=8cm,width=15cm]{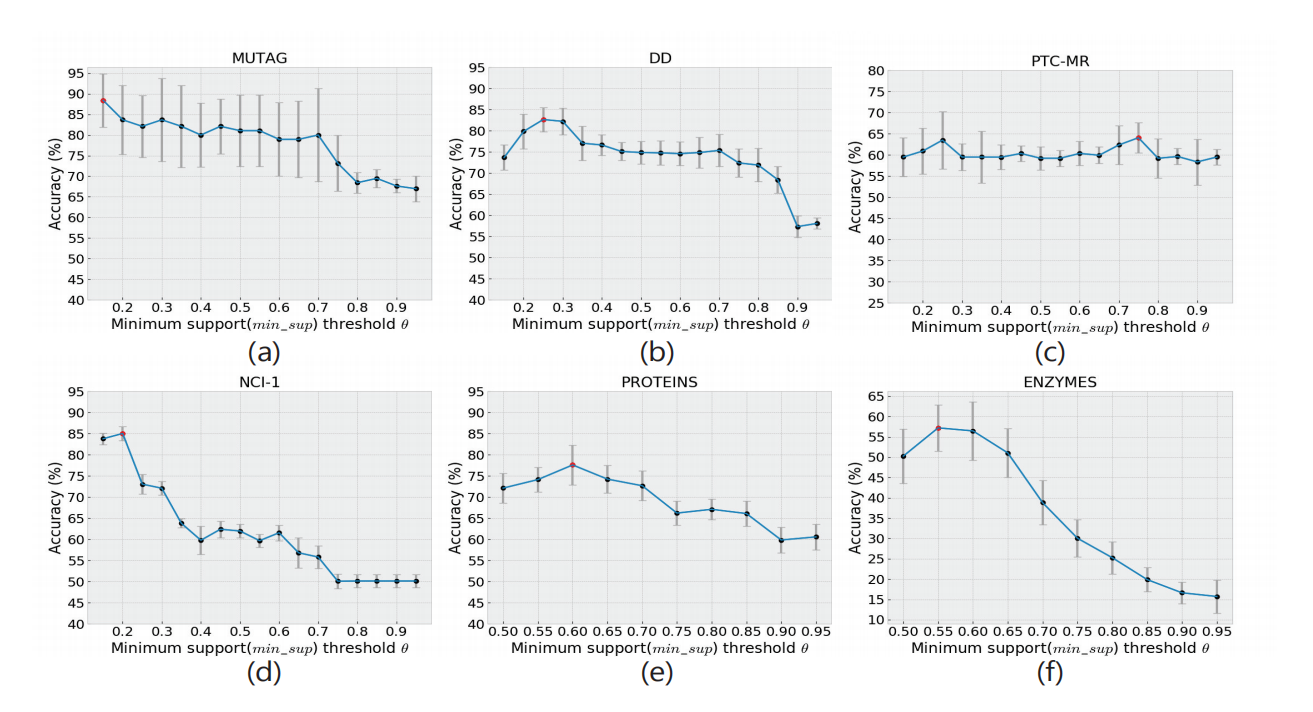}
\caption{Here we show parameter sensitivity of $min\_sup$ threshold $\theta$ in graph classification on MUTAG, DD, PTC-MR, NCI-1, PROTEINS and ENZYMES with fixed embedding dimension $dim = 128$. The best expressions marked by red points are used in our experiments and the $(\pm)$standard deviation of each result is indicated with gray error bar.}
\label{Figure-6}
\end{figure*}

\begin{figure*}[htb]
\centering
\includegraphics[height=8cm,width=15cm]{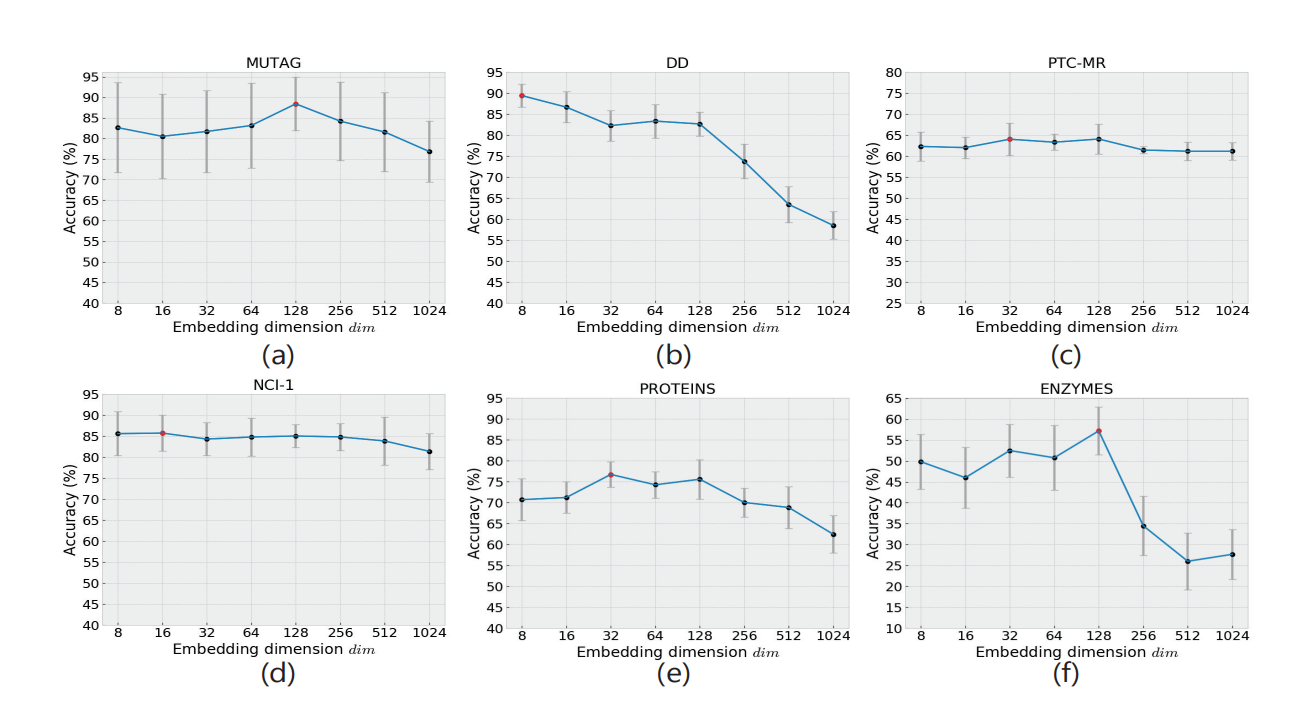}
\caption{Parameter sensitivity of embedding dimension $dim$ in graph classification on MUTAG, DD, PTC-MR, NCI-1, PROTEINS and ENZYMES with best $min\_sup$ threshold $\theta$ selected in Fig.5. The best expressions marked by red points are used in our experiments. The $(\pm)$standard deviation of each result is indicated with gray error bar.}
\label{Figure-7}
\end{figure*}

\subsection{Evaluation Metrics}
To evaluate the performance of GraphCSC, we randomly split the data into $10$ roughly equal-size parts and perform a $10$-fold cross validation on each dataset, in which $9$ folds for training and $1$ fold as validation for testing. This process is repeated $10$ times and an average accuracy is reported as prediction. Since we focus on graph embedding not on a classifier, we feed the embedding vectors to Support Vector Machine (SVM) with RBF kernel function with parameter $\sigma$ varing from the range $[10^{-4}, 10^{-3}, 10^{-2}, 10^{-1}, 1, 10]$.

\subsection{Parameter sensitivity}
Since we tend to prove that the performance of our model GraphCSC may have a gain on experiments results than method only considers skeletons, thus to explore how the hyperparameters in skeleton module affect tasks performace will not be our priority. For the need of brevity, the length $l$ is set as $10$ to generate a corpus of co-occurred anonymous random walks for all given datasets. To approximate actual distribution of anonymous random walks, we follow ~\cite{14_ivanov2018anonymous} to set the sampling hyperparameters $\varepsilon = 1$ and $\delta = 0.05$. In order to evaluate how the parameters sensitivity of $min\_sup$ threshold $\theta$ and embedding dimensions $dim$ affect the classification performance of GraphCSC on datasets, all parameters are assumed to be default except $\theta$ and $dim$.

We first conduct experiments with $dim=128$, then assess the classification accuracy as a function of $min\_sup$ threshold $\theta$ for different datasets. Best performance is indicated with red mark in Figure \ref{Figure-6}. Then in Figure \ref{Figure-7}, experiments examine the influence of varying the $dim$ from $[8, 16, 32, 64, 128, 256, 512, 1024]$ with the best $\theta$s obtained from Fig.5. The best accuracy is also signed in red as shown in each figure.

\subsection{Results and Discussion}
The average classification accuracy (standard deviation) over 10-fold cross-validation of GraphCSC (with $\theta$ and $dim$ that lead to optimal result from Figure \ref{Figure-6} and Figure \ref{Figure-7}, and baselines on seven real-world datasets are summarized in Table \ref{table-2-results}. From the results, it is evidently shown that GraphCSC is always the best in terms of performance on 6 datasets with exception on MUTAG, where GraphCSC gets second best result. More specifically, the proposed model achieves $3.41\%-30.60\%$ improvement over graph kernals based methods (WL, Deep WL, GK and Deep GK), and is competitive against graph embedding method (graph2vec) with $3.41\%-30.74\%$ gain in accuracy. However, supervised graph embedding method PSCN is more superior with 4.21\% higher in accuracy in MUTAG dataset classification. It is also obvious that GraphCSC outperforms AWE approach in every single dataset. This significantly demonstrates the effectiveness of the proposed model on classification tasks due to having access to skeleton and component features, which enable GraphCSC to get more complex structures information, while AWE processes only fixed size skeletons instead.

Figure \ref{Figure-6} presents the classification accuracy of increasing $min\_sup$ threshold $\theta$. In MUTAG and NCI-1, experiments get best performance when $\theta = 0.15$ and $0.2$ separately. Then results decrease slowly and maintain stable, after $\theta$ reaching $0.7$ (for MUTAG) and $0.6$ (for NCI-1), lines drop rapidly until getting to another flat states. A common phenomenon is that for DD, PROTEINS and ENZYMES, accuracy lines increase from the beginnings and decrease slowly after meeting the tops. Another observation is that in PTC-MR, performance has 2 closed top best results when $\theta = 0.25$ and $0.75$, while the result at $\theta = 0.75$ has only $1\%$ gain over that at $\theta = 0.25$. It is very interesting that $min\_sup$ threshold $\theta$ centers around a low value for most datasets. A possible explanation is that low $\theta$ will lead to more sufficient frequent subgraphs which will provide adequate complementary structure information for pattern just uses skeletons. As a consequence, this enforces model's discriminate power.

Figure \ref{Figure-7} examines the effects of embedding dimension hyperparameter $dim$. As we can see from the figure, in MUTAG and PROTEINS, there is a relatively gentle ﬁrst-increasing and then-decreasing accuracy line when $dim$ increases. We note that for DD, the performance line goes down slowly and maintains stable at around $83\%$ and then again falls quickly. Accuracy in ENZYMES changes roughly, indicating that classification result significantly suffers from embedding dimension $dim$. While it seems that $dim$ has no clearly effects on classification results for PTC-MR and NCI-1. The beat $dim$s for MUTAG and ENZYMES center at 128, but the $dim$ values tend to be small (usually no more than 32) in DD, PTC-MR, NCI-1 and PROTEINS, we can infer that small $dim$s would reflect more obvious features for these 4 datasets.

\section{Conclusion}
In this paper, we focus to study the problem of graph classification and propose GraphCSC, a new methodology that learns graph representation from horizontal and vertical perspectives to mine graph structures, i.e., skeletons and components. And our model is demonstrated to be superior to approach which utilizes only skeletons. Our method treats a graph as a document and different substructures or subgraphs in it as words by \textbf{NLP} framework \textbf{PV-DBOW}, thus the graph embedding GraphCSC learns integrate various size structures information. Several real-world graph classification tasks show that our model can achieve promising performances than a list of state-of-art baselines.

For future work, an interesting problem needs to be study further. We tend to fucus on more complex graph structures, for example, clusters. Since a cluster usually contains dozens or even hundreds of nodes and edges, there may show more valuable features which may reflect more implicit graph properties. We would also like to investigate graph embedding with mixed structures from low-order subgraphs to big clusters, and to verify whether this would be helpful for better results than GraphCSC.

\subsection*{Acknowledgements}
This work is supported by the Fundamental Research Funds for the Central Universities, the National Natural Science Foundation of China (No.11201019), the International Cooperation Project No.2010DFR00700, Fundamental Research of Civil Aircraft No. MJ-F-2012-04, the Beijing Natural Science Foundation (1192012, Z180005) and National Natural Science Foundation of China (No.62050132).

\bibliographystyle{unsrt}
\bibliography{bibfile.bib}

\end{document}